\documentclass[10pt,twocolumn,letterpaper]{article}

\usepackage{iccv}
\usepackage{times}
\usepackage{epsfig}
\usepackage{graphicx}
\usepackage{amsmath}
\usepackage{amssymb}
\usepackage{gensymb}
\usepackage{bm}
\usepackage{subfig}
\usepackage{multirow}
\usepackage{tabu}
\usepackage{makecell}
\usepackage{array}
\usepackage{tablefootnote}
\newcolumntype{?}{!{\vrule width 1pt}}

\usepackage[pagebackref=true,breaklinks=true,letterpaper=true,colorlinks,bookmarks=false]{hyperref}

\iccvfinalcopy 

\def\iccvPaperID{3169} 

\ificcvfinal\fi

\begin{document}

\newcommand{\john}[1]{\textcolor{red}{#1}}
\newcommand{\todo}[1]{\textcolor{red}{#1}}
\newcommand{\brandon}[1]{\textcolor{blue}{#1}}

\title{Black-Box Test-Time Shape REFINEment \\ for Single View 3D Reconstruction}

\author{Brandon Leung\\
UC San Diego\\
{\tt\small b7Leung@ucsd.edu}
\and
\and
Chih-Hui Ho\\
UC San Diego\\
{\tt\small chh279@ucsd.edu}
\and
Nuno Vasconcelos\\
UC San Diego\\
{\tt\small nvasconcelos@ucsd.edu}
}

\abovedisplayskip4pt
\abovedisplayshortskip=4pt
\belowdisplayskip=4pt
\belowdisplayshortskip=4pt

\maketitle
\ificcvfinal\thispagestyle{empty}\fi

\begin{abstract}
Much recent progress has been made in reconstructing the 3D shape of an object from an image of it, i.e. single view 3D reconstruction. However, it has been suggested that current methods simply adopt a ``nearest-neighbor''  strategy, instead of genuinely understanding the shape behind the input image. In this paper, we rigorously show that for many state of the art methods, this issue manifests as (1) inconsistencies between coarse reconstructions and input images, and (2) inability to generalize across domains. We thus propose REFINE, a postprocessing mesh refinement step that can be easily integrated into the pipeline of any black-box method in the literature. At test time, REFINE optimizes a network per mesh instance, to encourage consistency between the mesh and the given object view. This, along with a novel combination of regularizing losses, reduces the domain gap and achieves state of the art performance. We believe that this novel paradigm is an important step towards robust, accurate reconstructions,  remaining relevant as new reconstruction networks are introduced.
\end{abstract}

\vspace{-25pt}

\section{Introduction}

\begin{figure}
\begin{center}
\includegraphics[width=7cm]{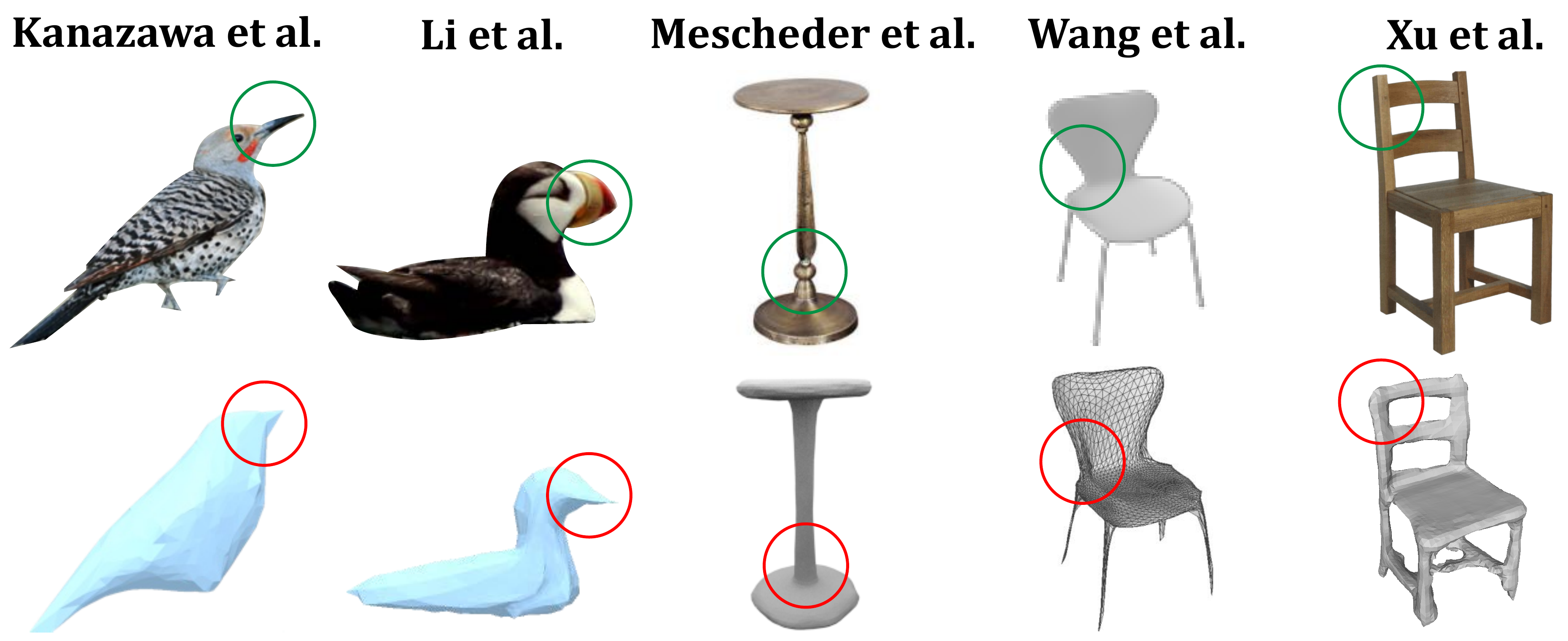} \\
\end{center}
   \caption{Important image details (circled in green) are frequently lost by state of the art 3D reconstruction methods (circled in red) \cite{kanazawa2018learning, li2020self, mescheder2019occupancy, wang2018pixel2mesh, xu2019disn}.}
\label{fig:rec_comparisions}
\end{figure}

Single view 3D reconstruction is the problem of reconstructing the 3D shape of an object from an image of it. Despite tremendous recent progress, the problem remains a significant challenge for computer vision. Figure \ref{fig:rec_comparisions} shows reconstructions by several state of the art methods. While the reconstructed shape (bottom row) reflects the category of the object in the image (top row), many details that determine fine-grained identity are lost. This happens even without ambiguity from self-occlusion, e.g. for object parts clearly visible in the image (circled in the figure). In contrast, unseen shape regions in the image tend to be reconstructed as well as visible ones. This suggests that current methods simply recognize the object, perform a ``nearest-neighbor'' search for a ``mean class shape''  memorized during training~\cite{tatarchenko2019single}, and make slight adjustments that are usually not enough to recover intricate shape details.

\begin{figure}
\begin{center}
\includegraphics[width=7.5cm]{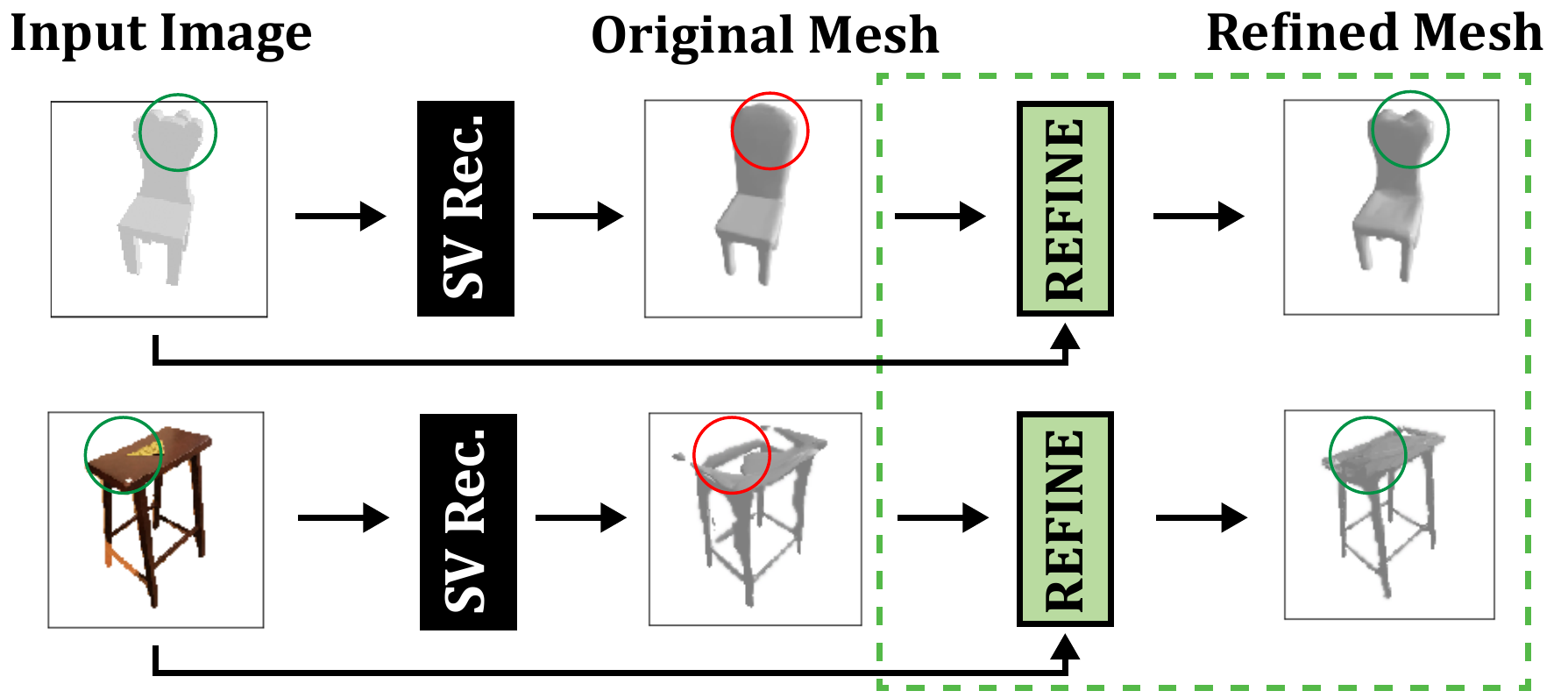} \\
\end{center}
   \caption{Black-box test-time shape refinement. Reconstructions by a network trained on ShapeNet are fed to our proposed {\it external} shape REFINEment network (in green) at test-time. REFINE improves reconstructions for images in both the training domain (top, image also from ShapeNet), and unknown test domains (bottom, image from Pix3D).}
\label{fig:domain_gap}
\end{figure}

This problem can be subtle when training and test distributions are similar, in which case nearest-neighbor produces decent reconstructions. However, it is magnified when there is a domain shift, as illustrated in Figure \ref{fig:domain_gap}. When a reconstruction network trained on ShapeNet \cite{chang2015shapenet} is applied to real images (Pix3D \cite{sun2018pix3d}), the reconstructed shape can have severe reconstruction failures. Distribution shift can often be mitigated with domain adaptation techniques ~\cite{sun2016deep, ganin2016domain, tzeng2017adversarial, long2017deep}. However, because they align entire probability distributions, these methods are likely to only help produce better mean class shape reconstructions. It is unclear that they will suffice to recover the lost shape details of Figure~\ref{fig:rec_comparisions}. They also require sizeable amounts of data from the target domain. A more ambitious but practical solution is to bridge shifts to unknown target domains by performing optimization at test time, in a self-supervised manner~\cite{sun2020test}. 

Recently,~\cite{remelli2020meshsdf} has shown that test time optimization can improve shape reconstructions, by introducing a differentiable explicit surface mesh representation for Deep Signed Distance Function (DeepSDF) \cite{park2019deepsdf} based reconstruction networks. This enabled the introduction of the MeshSDF algorithm which at test time fine-tunes a DeepSDF based network's weights to the target object using its silhouette. While test-time fine-tuning opens exciting possibilities to solve the problem of Figure~\ref{fig:rec_comparisions}, it has the limitations inherent to a white box technique. First, the original shape reconstruction network must be available and accessible for retraining. This is not always easy, given the complexity of many state of the art reconstruction procedures. Second, the test-time optimization is specific to that network. For example, it is not trivial to extend the method of~\cite{remelli2020meshsdf} to approaches like Mesh R-CNN~\cite{gkioxari2019mesh}  which utilizes an intermediate voxel representation, nor Atlasnet~\cite{groueix2018papier}, which represents meshes using atlas surface elements. \looseness=-1 

This suggests the need for {\it black-box\/} single view reconstruction refinement, which would not suffer from these problems. Unfortunately, no black-box refinement method currently exists. Thus, in this work we design a dedicated refinement network {\it external\/} to the reconstruction model, which refines the mesh shape produced by the latter, a posteriori as illustrated in Figure \ref{fig:domain_gap}. In contrast to network fine-tuning, black-box refinement has the advantage of being model agnostic and applicable even to meshes produced by third-party networks (e.g. ~\cite{choy20163d, tatarchenko2017octree, mescheder2019occupancy, groueix2018papier, kanazawa2018learning}) unavailable at test time. Furthermore, it allows the joint development of network architectures and loss functions tailored to solving the problem of Figure~\ref{fig:rec_comparisions} at test time. This is important because, when compared  to \textit{reconstruction networks} trained with large datasets, test-time \textit{refinement networks} must be much more efficient and less prone to overfitting. 

In particular, our proposed novel black-box test-time shape refinement procedure for single view reconstruction follows the test-time formulation of~\cite{remelli2020meshsdf}, i.e. seeks the test-time shape refinement that best matches an object's silhouette, but abstracts this refinement from the original shape reconstruction. Rather than simply finetuning the reconstruction network, we introduce a REFINE (a recursive acronym for \textbf{REF}ine \textbf{IN}stances at \textbf{E}valuation) procedure, implemented with a {\it postprocessing\/} network whose parameters are optimized at test time using the given reconstructed shape and object silhouette. The REFINE optimization is performed at instance level, i.e. each instance is refined independently with re-initialized parameters. It leverages a novel combination of loss functions, encouraging both silhouette consistency and confidence-based mesh symmetry, to produce mesh displacements. Using a variety of metrics, it is systematically shown that reconstruction quality of existing networks degrades when training and testing distributions are different. The degradation is studied on (1) different renderings of a synthetic dataset and (2) real world datasets. It is then shown that REFINE improves the shapes reconstructed by various reconstruction networks, on-the-fly,  without relying on prior target distribution knowledge. This holds even when there are large domain gaps (see Figure \ref{fig:domain_gap}), and outperforms the state of the art test-time finetuning method of~\cite{remelli2020meshsdf} even for DeepSDF reconstructions.

Overall, this work makes three main contributions. The first is to characterize the inconsistency between input image and mesh reconstructed by existing networks. This shows that these networks (1) produce coarse, class-averaged, shapes and (2) are unable to tackle distribution shift between training and test data. Second, a new black-box test-time shape REFINEment framework, based on instance-level test-time optimization, is proposed to overcome the problem. Finally, extensive experiments demonstrate that REFINE achieves state of the art performance for test-time shape refinement, leading to significantly more accurate shapes than those synthesized by current reconstruction networks, especially under distribution shift.

\section{Related Work}

In this section, we briefly review the related literature in single view reconstruction and test-time optimization.

\paragraph{Single View 3D Reconstruction.}
While many single view 3D reconstruction methods have been proposed, they all suffer from the inconsistencies of Figure~\ref{fig:rec_comparisions}, and can benefit from REFINE. The main 3D output modalities are voxels, pointclouds, and meshes. Voxel based methods typically encode an image into a latent vector, which is decoded into a 3D voxel grid with upsampling 3D transposed convolutions \cite{xie2019pix2vox, choy20163d}. Octrees can enable higher voxel resolution \cite{tatarchenko2017octree, wang2017cnn}. Pointclouds have been explored as an alternative to voxels \cite{fan2017point, lin2017learning} but usually require voxel or mesh conversion for use by downstream tasks. Among mesh based methods, some learn to displace vertices on a sphere \cite{kato2018neural, wang2018pixel2mesh} or a mean shape \cite{kanazawa2018learning} to produce the output. Current state of the art methods rely on an intermediate implicit function representation to describe shape \cite{mescheder2019occupancy, xu2019disn, park2019deepsdf, genova2020local, niemeyer2020differentiable}, mapped into a mesh by marching cubes~\cite{lorensen1987marching}.

Methods also vary by their level of supervision. Most are fully supervised, requiring a large dataset of 3D shapes such as ShapeNet \cite{chang2015shapenet}. Recently, weakly-supervised methods have also been introduced, using semantic keypoints \cite{kanazawa2018learning} or 2.5D sketches \cite{wu2017marrnet} as supervision. Alternatively, \cite{li2020self} has proposed a fully unsupervised method, combining part segmentation and differentiable rendering. Few-shot is considered in ~\cite{wallace2019few, michalkiewicz2020few}  where classes have limited training data. Domain adaptation was explored in \cite{pinheiro2019domain}, which assumes access to data from a known target domain.

Despite progress in single view 3D reconstruction, questions arise on what is actually being learned. In particular, \cite{tatarchenko2019single} shows that simple nearest-neighbor model retrieval can beat state of the art reconstruction methods. This raises concerns that current methods bypass genuine reconstruction, simply combining image recognition and shape memorization. Such memorization is consistent with Figures \ref{fig:rec_comparisions} and \ref{fig:domain_gap}, leading to suboptimal reconstructions and inability to generalize across domains. It is likely a consequence of learning the reconstruction network over a training set of many instances from the same class. In contrast, REFINE uses test-time optimization to refine a single shape, encouraging consistency with a single silhouette. This prevents memorization, directly addressing the concerns of \cite{tatarchenko2019single}. It also makes REFINE complementary to reconstruction methods and applicable as a postprocessing stage to any of them. 

\paragraph{Test-Time Optimization.}
Test-time optimization usually exploits inherent structure of the data in a self-supervised manner, as no ground truth labels are available. For example, \cite{sun2020test} leverages an auxiliary self-supervised rotation angle prediction task at test time to reduce domain shift in object classification. The same goal is achieved in \cite{wang2020fully} by test-time entropy minimization. Meanwhile, \cite{tung2017self} uses self-supervision at test time to improve human motion capture. Additionally, interactive user feedback serves to dynamically optimize segmentation predictions \cite{sakinis2019interactive, sofiiuk2020f, jang2019interactive}.

\begin{figure*}
\begin{center}
\includegraphics[width=16.5cm]{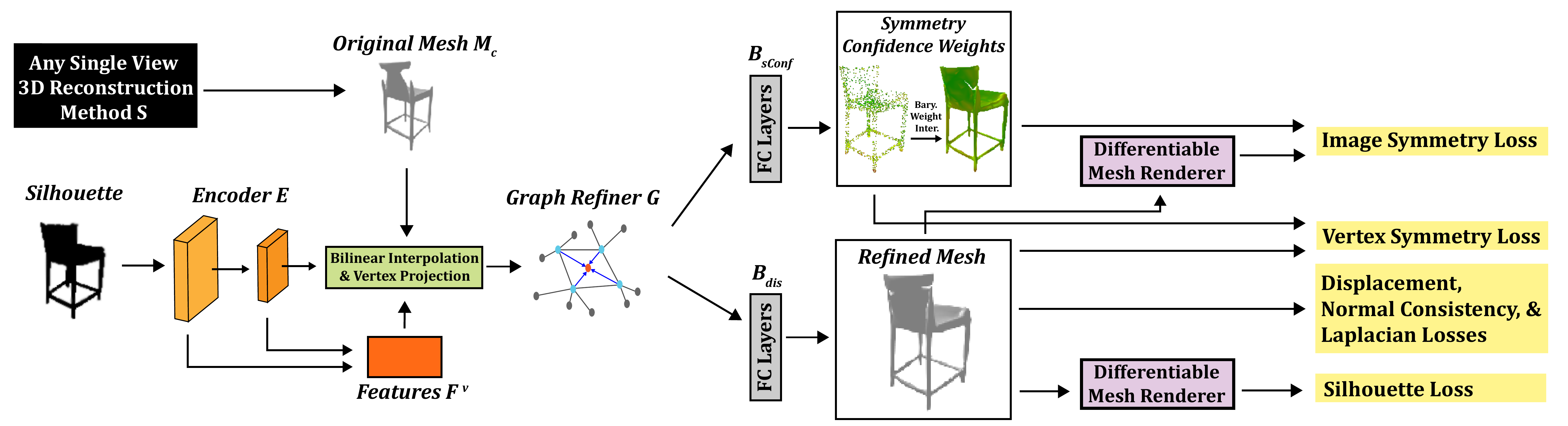} \\
\end{center}
   \caption{The REFINE network architecture. Given an original mesh reconstruction with missing details, the network outputs vertex translations needed to make the refined mesh consistent with an input image. It consists of an silhouette feature map encoder, whose outputs create a shape graph. This is refined by graph convolutions, and two fully connected branches which output the refined mesh and symmetry confidence weights. The latter enforces optimized 3D symmetry constraints. Several losses aid to avoid overfitting to the input viewpoint. Optimization is performed over single examples, at test time.}
\label{fig:deform_net}
\end{figure*}

\paragraph{Test-Time Shape Refinement.} 
Test-time shape refinement requires a postprocessing procedure to improve the accuracy of meshes produced by a reconstruction network. Most previous approaches are white-box methods, i.e. they are specific to a particular model (or class of models) and require access to the internal workings of the model. Examples include methods that exploit temporal consistency in videos, akin to multi-view 3D reconstruction \cite{li2020onlineAda, lin2019photometric}. \cite{li2020onlineAda} requires the  unsupervised part-based video reconstruction architecture proposed by the authors and \cite{zuffi2019three} optimizes over a shape space specific to their architecture for zebra images. Among white-box methods, the approach closest to REFINE is that of \cite{remelli2020meshsdf}, which finetunes the weights of the reconstruction network at test time, to better match the object silhouette. But even this method is specific to sign distance function (DeepSDF \cite{park2019deepsdf}) networks. By instead adopting the black-box paradigm, where the mesh refinement step is intentionally decoupled from the reconstruction process, REFINE is capable of learning \textit{vertex-based deformations} for a mesh generated by {\it any reconstruction architecture\/}. Our experiments show that it can be effectively applied to improve the reconstruction performance of many networks and achieves state of the art results for test-time shape refinement, even outperforming \cite{remelli2020meshsdf} for DeepSDF networks. In summary, unlike prior approaches, REFINE is a black-box technique that can be universally applied to improve reconstruction accuracy, a posteriori.


\begin{table}
\begin{center}
\scriptsize
\setlength\tabcolsep{1pt}
\begin{tabular}{|c|c|c|c|c|c|}
\hline
\textbf{Method} & REFINE & OccNet \cite{mescheder2019occupancy} & MeshSDF \cite{park2019deepsdf} & Pix2Mesh \cite{wang2018pixel2mesh} & AtlasNet \cite{groueix2018papier}\\
\hline
\textbf{Params. (Mil.)} & 0.9 & 12.7 & 13.2 & 18.8  & 20.3  \\
\hline
\end{tabular}
\end{center}
\caption{Parameter size comparisons between the REFINE network and popular single view reconstruction models.}
\label{tab:params_compare}
\end{table}

\section{Black-Box Test-Time Shape Refinement}
In this section we introduce REFINE, detailing its neural network architecture, losses, and training procedure.

\subsection{Formulation and Inputs}
Single view 3D reconstruction methods reconstruct a 3D object shape from a single image of the object. An RGB image $x \in \mathbb{R}^{W\times H \times 3}$ of width $W$ and height $H$ is mapped to a mesh $M$ by a reconstruction network 
\begin{equation}
S:\mathbb{R}^{W\times H \times 3}\rightarrow \mathcal{M} \in \mathcal{V}\times \mathcal{E},
\end{equation}
i.e., $S(x) = M = (V, E)$ where $V \in \mathcal{V}\subset\mathbb{R}^{N\times3}$ is a set of vertices and $E \in \mathcal{E}\subset\mathbb{B}^{N \choose 2}$ a set of edges. $\mathbb{B}$ is a boolean domain specifying mesh connectivity. $S(x)$ is usually a coarse shape estimate, whose details do not match the input image, as shown in Figure \ref{fig:rec_comparisions}. Performance further degrades when $x$ is sampled from an image distribution different from that used during training \cite{pinheiro2019domain}.

It was proposed in \cite{remelli2020meshsdf} to address this issue by optimizing the parameters of $S$ on-the-fly during inference, given a coarsely reconstructed mesh $S(x) = M_c = (V_c, E_c)\in\mathcal{M}$, an object silhouette $x_{s}$, and the camera pose $p$. We refer to this problem as {\it test-time shape refinement\/} (TTSR). We investigate an alternative class of solutions to the TTSR problem,
which abstracts shape refinement from the reconstruction network $S$. These {\it black-box} solutions implement a refinement mapping  with a dedicated refinement network $R$ external to $S$. The network $R$ is trained at test-time, so that $R \circ S(x)$ is a 3D mesh that more accurately approximates the object shape, as illustrated in Figure~\ref{fig:domain_gap}. We denote the approach as REFINE and $R$ as the REFINEment network. In this formulation, $R$ predicts a set of 3D displacements $V_{dis} \in \mathbb{R}^{N\times3}$  for the vertices in $V_c$. These are used to compute the REFINEd mesh $M_r=(V_r,E_r)=(V_c+V_{dis}, E_c)$ whose render best matches the silhouette $x_{s}$. This set of displacements is complemented by a set of symmetry confidence scores $V_{sConf} \in [0,1]^{N\times1}$, which regularizes $V_{dis}$ through a symmetry prior, as elaborated upon in Section \ref{sec:loss}.

Several advantages derive from the abstraction of refinement from reconstruction. First, this makes REFINE a black-box technique, applicable to any network $S$. In fact, the network does not even have to be available, only the mesh $S(x)$, which gives REFINE a great deal of flexibility. For example, while MeshSDF can only be used with DeepSDF networks, REFINE is applicable even to voxel and pointcloud reconstruction methods, by using mesh conversions \cite{kazhdan2013screened, kazhdan2006poisson, calakli2011ssd, bernardini1999ball}. This property is important, as different methods are better suited for different downstream applications. For example, implicit methods \cite{mescheder2019occupancy, park2019deepsdf} tend to produce the best reconstructions but can have slow inference \cite{park2019deepsdf}. Meanwhile, AtlasNet \cite{groueix2018papier} is less accurate but much more efficient, and inherently provides a parametric patch representation useful for downstream applications like shape correspondence. A second advantage of black-box refinement is that because the refinement network $R$ and loss functions used to train it are independent of the reconstruction network $S$, they can be specialized to the test-time shape refinement goal. This is important because special considerations must usually be taken to regularize $R$ when compared to $S$, since test time training is based on a single mesh instance, rather than a large dataset. To avoid overfitting,  we design $R$ to be much smaller than $S$. As shown in Table \ref{tab:params_compare}, the REFINE network is at least 10 times smaller than most currently popular reconstruction networks. We also propose several novel loss functions, tailored for test-time training, to regularize $R$.

\subsection{Architecture} \label{sec:arch}

Figure \ref{fig:deform_net} summarizes the architecture of the REFINE network. This is a combination of an encoder $E$ and a graph refiner $G$ 
followed by 2 branches $B_{dis}$ and $B_{sConf}$, which predict the vertex displacements and vertex confidence scores respectively. The encoder module $E$ contains $L$ neural network layers of parameters $\{\theta_i\}_{i=1}^L$, takes silhouette $x_{s}$ as input, and outputs a set of $L$ feature maps  $F(x_{s};\Theta_l=\{\theta_j\}_{j=1}^l) \in \mathcal{R}^{W_l\times H_l \times C_l}$, of width $W_l$, height $H_l$ and $C_l$ channels. In our implementation, $E$ is based on ResNet \cite{he2016deep}; $L$ is set to 2, where $C_1=256$ and $C_2=512$.

Given feature map $F(x_{s};\Theta_l),$ the feature vector $f_l^v$ corresponding to a vertex $v$ in $M_c$ is computed by projecting the vertex position onto the feature map \cite{gkioxari2019mesh,wang2018pixel2mesh},
\begin{equation}
    f_l^v = Proj(v;F(x_s;\Theta_l),p) \in \mathcal{R}^{C_l},
\end{equation}
where $p$ is the camera viewpoint and $Proj$ a perspective projection with bilinear interpolation. Vertices are represented at different resolutions, by concatenating the feature vectors of different layers into $F^v=(f_1^v, \dots , f_L^v)^T$. The set $\{F^v\}_{v=1}^{N}$ of concatenated feature vectors extracted from all vertices is then processed by a graph convolution \cite{kipf2016semi} refiner $G$, of parameters $\phi$, to produce an improved set of feature vectors
$\{H^v\}_{v=1}^{N} = G(\{F^v\}_{v=1}^{N};\phi)$. Finally, this set is mapped into the displacement vector $V_{dis}$ 
\begin{equation}
    V_{dis} = B_{dis}(\{H^v\}_{v=1}^{N};\psi_{dis}),
\end{equation}
by a fully connected branch $B_{dis}$ of parameters $\psi_{dis}$ and into the confidence vector 
\begin{equation}
   V_{sConf} =
   B_{sConf}(\{H^v\}_{v=1}^{N};\phi);\psi_{sConf})
\end{equation}
by a fully connected branch $B_{sConf}$ of parameters $\psi_{sConf}$. Overall, the REFINE network implements the mapping 
\begin{equation}
    R(x_s,M_c;\{\theta_i\}, \phi, \psi_{dis}, \psi_{sConf},p) = \{V_{dis}, V_{sConf}\}. \label{eq:refine_arch}
\end{equation}
\vspace{+4pt}
\vspace{-16pt}
\subsection{Optimization} \label{sec:loss}

The REFINE optimization combines popular reconstruction losses with novel losses tailored for test-time shape refinement. In what follows we use $M^p$ to denote a differentiable renderer \cite{kato2018neural, liu2019soft} that maps mesh $M\in\mathcal{M}$ into its image captured by a camera of parameters $p$. We also define sets $V^s_r$, $V^s_{dis}$, and $V^s_{sConf}$ of size $N$, constructed with the rows of $V_r$, $V_{dis}$, and $V_{sConf}$ respectively. \\ 
\noindent\textbf{Silhouette Loss:} Penalizes shape and silhouette mismatch
\begin{equation}
    L_{Sil} = L_{BCE}(x_s, \gamma(M_r^{p})), \label{eq:sil}
\end{equation}
where $\gamma(M_r^{p})$ is the silhouette of the rendered shape,
using the 2D binary cross entropy loss
\begin{equation}
    L_{BCE}(a,b)=\sum_{ij}a_{ij}\log(b_{ij})+(1-a_{ij})\log(1-b_{ij}).
\end{equation}
\noindent\textbf{Displacement Loss:} Discourages overly large vertex deformations, with
\begin{equation}
    L_{Dis} = \sum_{v_i \in V^s_{dis}} ||v_i||_2^2.
\end{equation}
\\
\noindent\textbf{Normal Consistency \& Laplacian Losses:} $L_{Nc}$ and $L_{Lp}$ are widely used \cite{wang2018pixel2mesh, desbrun1999implicit} to encourage mesh smoothness.\\

\begin{figure}
\begin{center}
\includegraphics[width=4.5cm]{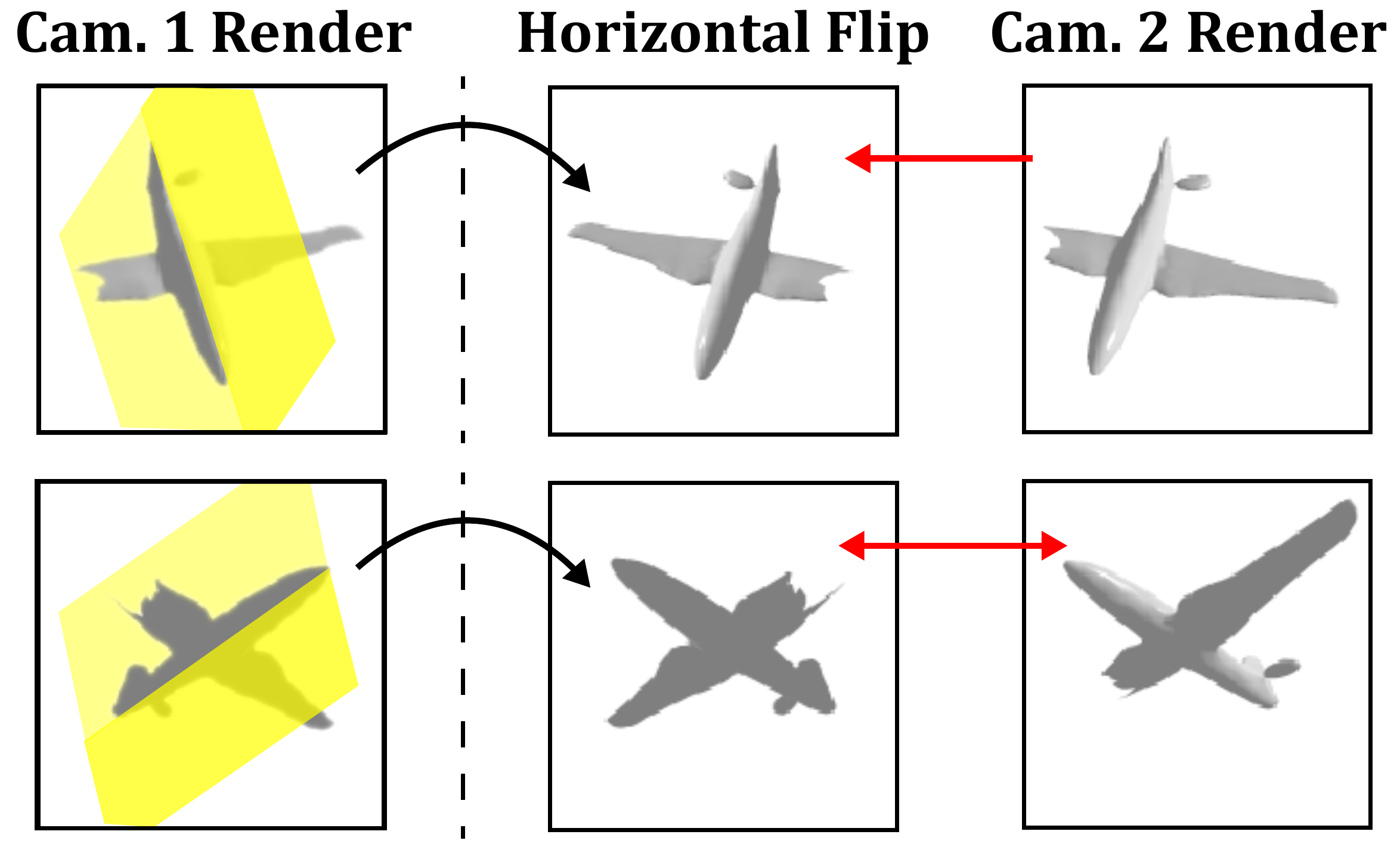} \\
\end{center}
   \caption{To enforce symmetry, a mesh is differentiably rendered by two cameras, where the viewpoint of camera 2 is obtained by reflecting that of camera 1 about the mesh's plane of symmetry (yellow). The second render is compared to the horizontal flip of the first.}
\label{fig:img_sym_loss}
\end{figure}

A second set of losses is introduced to avoid overfitting during test-time refinement. This leverages the prior that many real world objects are bilaterally symmetric  \cite{liu2010computational} about a reflection plane $\mathcal{Z}$. These losses enforce two constraints. The first is that the object vertices should be symmetric. The second is that the image projections of the shape should reflect that symmetry. The latter is enforced by the procedure of Figure \ref{fig:img_sym_loss}. For a given render, under camera 1, the reflected camera, i.e. the camera whose viewpoint is reflected about $\mathcal{Z}$, is first found. The object is then rendered under this camera viewpoint and the resulting render compared to the horizontal flip of the render under camera 1. All reflections are implemented with the transformation $T = I - 2 \vec{n}\vec{n}^\intercal$, where $\vec{n} \in \mathbb{R}^3$ is the unit normal vector of $\mathcal{Z}$. While there are methods to predict planes of object symmetry \cite{gao2019prs, zhou2020learning}, we found them to be unnecessary, since most reconstruction methods output semantically aligned meshes for objects of the same class. In general, the objects are aligned so that $\mathcal{Z}$ is the vertical plane with $\vec{n} = [0,0,1]^\intercal$. We adopt this convention in all our experiments.

While we have found this symmetry prior to be helpful for many objects, not all objects are symmetric or exactly symmetric. For example, an object can be almost symmetric (e.g. an airplane missing part of one wing). To prevent the symmetry prior from overwhelming (\ref{eq:sil}) when this is the case, we introduce a confidence score $\sigma_i$ per vertex $v_i$. These confidence scores are learned during the REFINE optimization, enabling local deviations from the global symmetry constraint when appropriate. The two symmetry losses are defined as follows.

\begin{figure}
\begin{center}
\includegraphics[width=8cm]{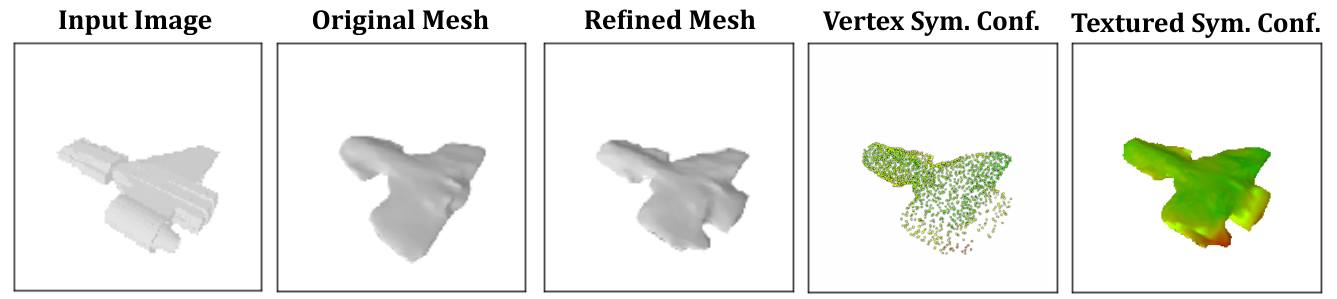} \\
    \end{center}
   \caption{Left to right: image, original mesh, REFINEd mesh, and vertex confidence weights (shown as points or colors on the REFINEd mesh). Red shades indicate lower confidence; green higher. Note how the confidence weights relax the symmetry prior on asymmetric object parts.}
\label{fig:asym}
\end{figure}

\noindent\textbf{Vertex-Based Symmetry Loss:} Encourages symmetric mesh vertices according to  
\begin{equation}
L_{Vsym} = \frac{1}{N} \sum^N_{i=1} \sigma_i \min_{v_j\in V^s_r} \left\Vert T v_i - v_j \right\Vert_2^2 + \lambda_{SymB}\ln\left(\frac{1}{\sigma_i}\right), \label{eq:vertex_sym}
\end{equation}
where $v_i \in V_r^s$ are the mesh vertices and $\sigma_i \in V^s_{sConf}$ the associated symmetry confidence scores.
The first term penalizes distances between each vertex and its nearest neighbor upon reflection about $\mathcal{Z}$. This is weighted by the learned confidence score $\sigma_i$, which is low for vertices that should be asymmetric based on the signal given by its silhouette.  The second term penalizes small confidence scores, preventing trivial solutions. Trade-offs between the two terms is controlled by hyperparameter  $\lambda_{SymB} \in [0,\infty)$. Shown in Figure \ref{fig:asym}, scores $\sigma_i$ are large except in areas of clear asymmetry.

\begin{figure}
\begin{center}
\includegraphics[width=7.5cm]{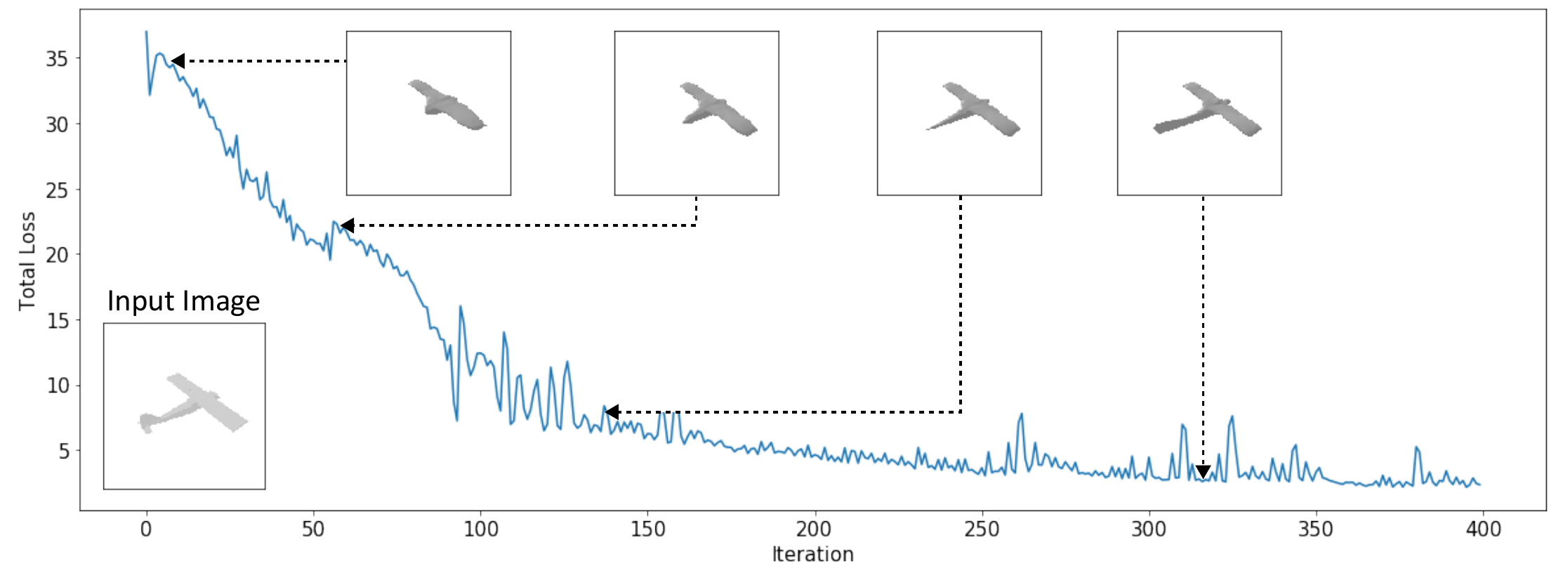} \\
\end{center}
   \caption{As the REFINEment optimization proceeds, 3D shape becomes more accurate.}
\label{fig:refine_progress}
\end{figure}

\noindent\textbf{Render-Based Image Symmetry Loss:} Encourages image projections that reflect object symmetry. Given $m$ camera viewpoints $P_{Isym}=\{p_1,...,p_m\}$, $T$ is used to obtain rendered pairs from symmetric camera viewpoints $\{(M_r^{p_1}, M_r^{Tp_1}), ..., (M_r^{p_m}, M_r^{Tp_m})\}$, as shown in the rows of Figure \ref{fig:img_sym_loss}. The loss is defined as
\begin{equation}
    \begin{aligned}
        L_{Isym} ={} \frac{1}{m} \sum^m_{i=1} \sum_{j,k} & \Big[ \sigma_{j,k}|| \gamma(h(M_r^{p_i}))_{j,k} - \gamma(M_r^{Tp_i})_{j,k} ||_2^2 \\
        & + \lambda_{SymB}\ln\left(\dfrac{1}{\sigma_{j,k}}\right) \Big],   
    \end{aligned}\label{eq:render_sym}
\end{equation}
where $h(\cdot)$ is image horizontal flip and $j,k$ are image coordinates. Symmetry is enforced by minimizing the distance between the horizontal flip of each render $M_r^{p_i}$ and the render $M_r^{Tp_i}$ at the symmetrical camera viewpoint. This is akin to comparing a ``virtual image'' of what the mesh should symmetrically look like. Pixel-based confidence scores $\sigma_{j,k}$ are used as in (\ref{eq:vertex_sym}). However, they are not relearned, but derived from the vertex confidences $\sigma_i, i \in \mathcal{V}_{j,k}$, of (\ref{eq:vertex_sym}) by barycentric interpolation on the mesh faces, where $\mathcal{V}_{j,k}$ are mesh face vertices projected into pixel $j,k$.\\
\noindent\textbf{Overall Loss:} REFINE is trained with a weighed combination of the six losses
\begin{equation} \label{eq:total_loss}
\begin{aligned}
        L_{total} ={} & \lambda_{Sil}L_{Sil} + \lambda_{Isym}L_{Isym} + \lambda_{Vsym}L_{Vsym} \\ 
              & + \lambda_{Dis}L_{Dis} + \lambda_{Nc}L_{Nc} + \lambda_{Lp}L_{Lp}.
\end{aligned}
\end{equation}
$L_{Sil}$ is the main driving factor to ensure input silhouette consistency, while other losses serve as regularizers to prevent overfitting. Figure \ref{fig:refine_progress} shows that REFINEd shape quality tracks the evolution of this loss, for an airplane whose body has been truncated in the original reconstruction. As the REFINE loss steadily decreases, the airplane mesh progressively becomes more faithful to the input image; this is seen in the elongated body and corrected wing shape.

\begin{figure}
\begin{center}
\includegraphics[width=7cm]{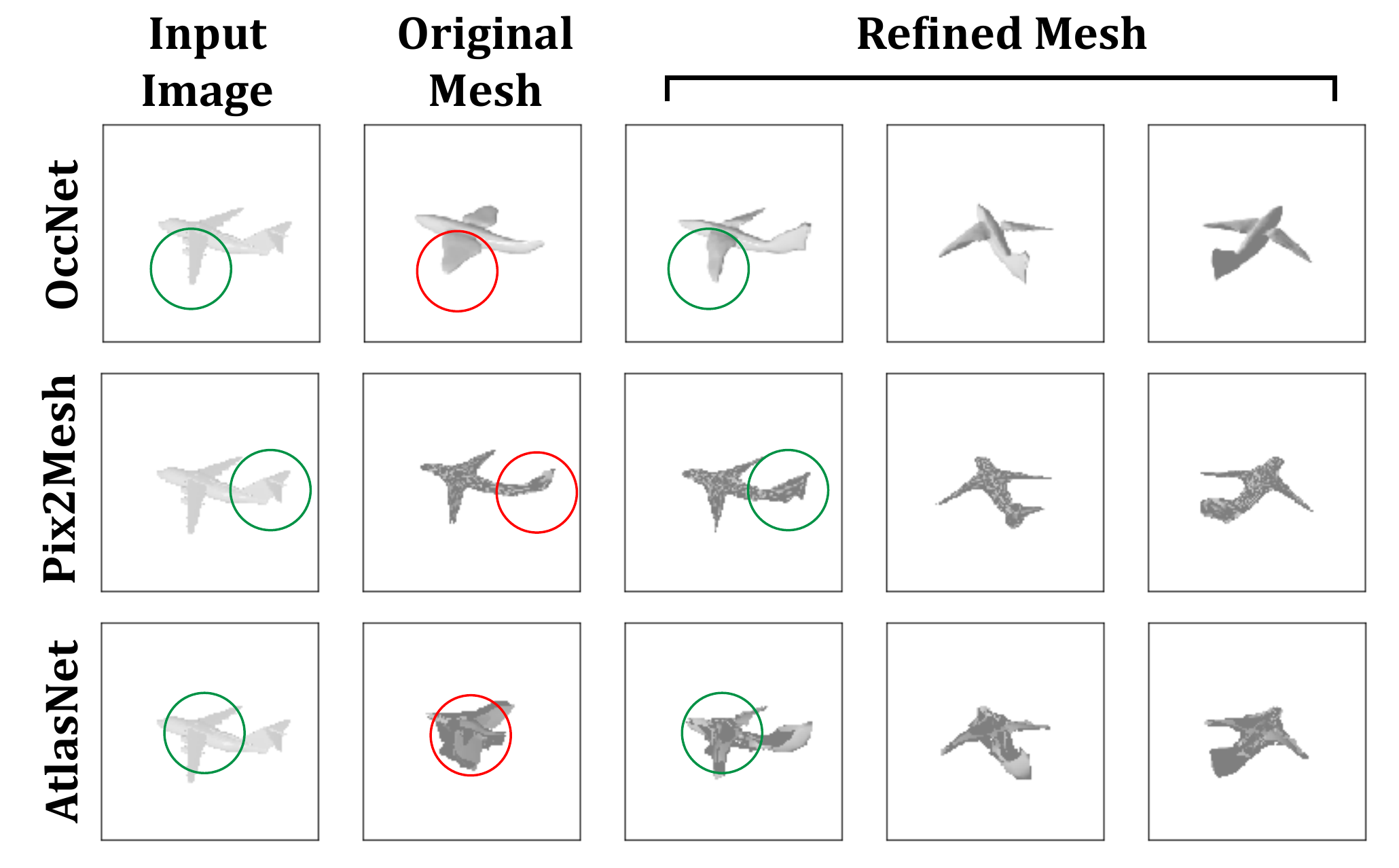} \\
\end{center}
   \caption{An airplane reconstructed by three different methods \cite{mescheder2019occupancy, wang2018pixel2mesh, groueix2018papier}. Since the methods differ greatly, they exhibit very different failure cases and artifacts. Nevertheless, REFINE improves all reconstructions.}
\label{fig:method_compare}
\end{figure}

\subsection{Implementation Details}
Ideally, test-time shape refinement postprocessing should support any mesh and be fast. REFINE intrinsically satisfies the first requisite, since it is black-box, class-agnostic, and allows variable number of vertices per mesh. To prevent the imprecise shapes of Figure \ref{fig:rec_comparisions}, it optimizes a single instance at a time, starting from a network of random weights. Optimization from scratch converges in relatively few iterations, approximately 400 (i.e. 400 forward and backward passes). This requires about 90 seconds on a GTX 1080Ti GPU. Moreover, because instances are treated independently, the refinement is trivially parallelizable. Since 4 instances fit on a GPU, a two GPU server trivially achieves a per-instance refinement time of $90/(4*2)\approx 11$ seconds, which is effective in terms of the second requisite. 

Several details of our implementation are worth noting. In all experiments we used $P_{Isym}$ of 6 viewpoints, with azimuths in $\{15\degree, 45\degree, 75\degree\}$ and elevations in $\{-45\degree, 45\degree\}$. The learning rate is 0.00007, $\lambda_{Sil}=10$, $\lambda_{Isym}=80$, $\lambda_{Vsym}=20$, $\lambda_{SymB}=0.0005$ $\lambda_{Dis}=100$, $\lambda_{Nc}=10$, and $\lambda_{Lp}=10$. More details are given in the supplementary.

\begin{figure}
\begin{center}
\includegraphics[width=6.5cm]{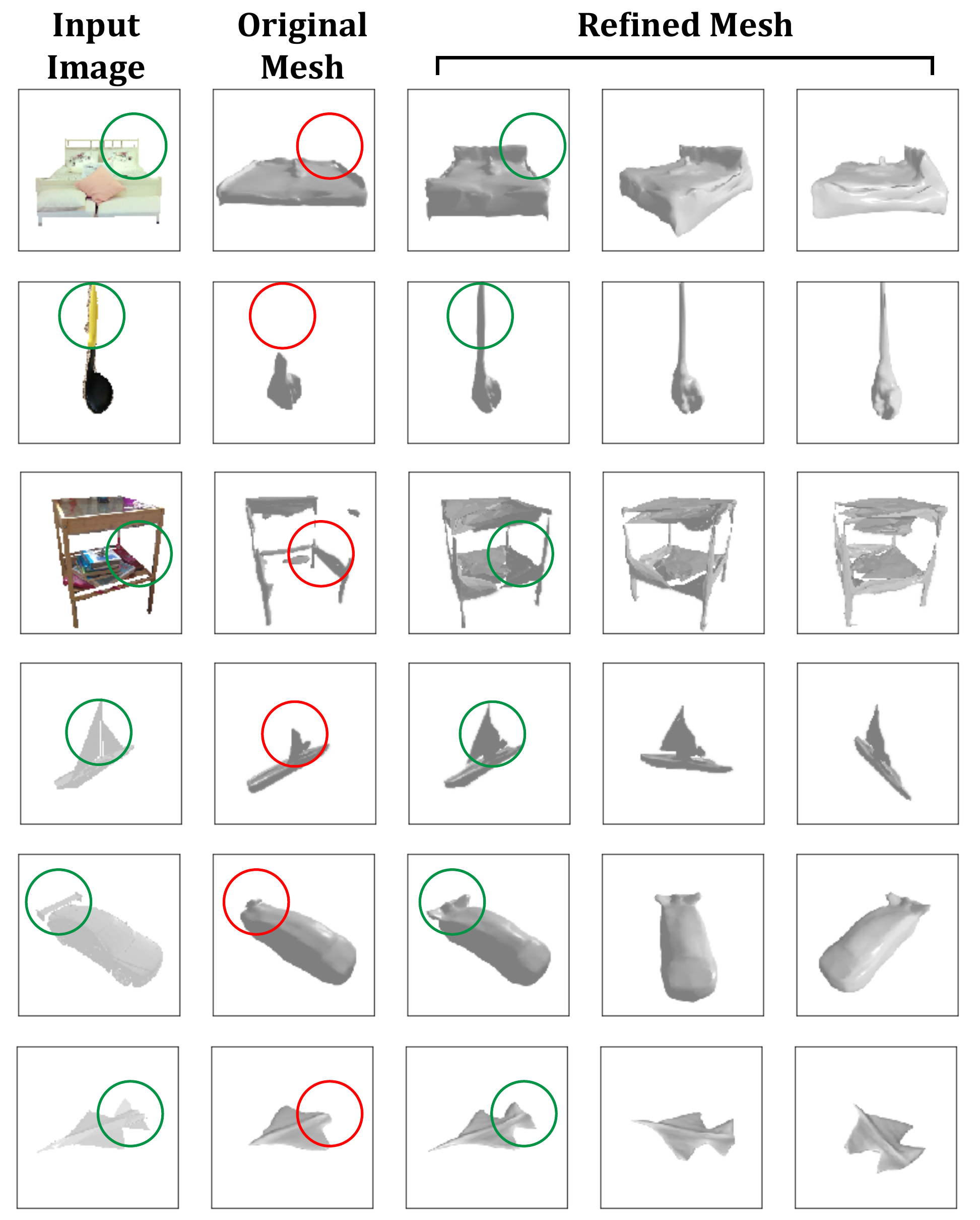} \\
\end{center}
   \caption{Mesh REFINEment examples for Pix3D (top three rows) and ShapeNet (bottom three) images. REFINE is capable of correcting small details (e.g. airplane nose) as well as generating entirely new parts (e.g. the rear wing).}
\label{fig:qual_examples}
\end{figure}

\section{Experiments} \label{sec:exp}
In this section, we discuss several experiments performed to validate the effectiveness of REFINE.

\subsection{Experimental Setup} \label{sec:setup}

\textbf{Metrics:} To evaluate REFINEment performance, the original mesh is first reconstructed by a baseline single view reconstruction method, and the reconstruction accuracy is measured. REFINE is then applied to the mesh and its accuracy is measured again. Several popular metrics \cite{tatarchenko2019single} are used to measure 3D accuracy: EMD, $l2$ Chamfer Distance, F-Score, and 3D Volumetric IoU. Lower is better for EMD and Chamfer distance, while higher is better for IoU and F-Score; for details please refer to the supplementary.


\textbf{Datasets:} Four datasets are considered, to rigorously study domain shift. All baseline models are trained on the synthetic \textit{ShapeNet} dataset \cite{chang2015shapenet}, with images rendered by \cite{choy20163d} using Blender~\cite{blender}. We also re-rendered the 3D models in the test set of \cite{choy20163d} using Pytorch3D \cite{ravi2020pytorch3d}. This second dataset, called \textit{RerenderedShapeNet} is designed to create a domain gap due to significant differences in shading, viewpoint, and lighting. The third dataset is motivated by our observation that about  97\% of ShapeNet is symmetrical, in the sense that each mesh has a symmetry loss $L_{Isym}<0.01$ for $\lambda_{SymB}=1$ and $\sigma_{j,k}=1$. To ablate how asymmetry affects reconstruction quality, we introduce a subset of RerenderedShapeNet, denoted as {\it ShapeNetAsym\/}, containing 1259 asymmetric meshes. Finally, we use the \textit{Pix3D} dataset \cite{sun2018pix3d}, which contains real images and their ground truth meshes, to test for large domain shifts. Hyperparameters were tuned with a small portion of RerenderedShapeNet, disjoint from the test set. \looseness=-1
\begin{table}
\begin{center}
\scriptsize
\setlength\tabcolsep{3pt}
\begin{tabular}{|c|c|c|c|c|c|}
\hline
Configuration & EMD$\downarrow$ & CD-$l_2 \downarrow$ & F-Score$\uparrow$ & Vol. IoU$\uparrow$ & 2D IoU \\
\hline\hline
OccNet \cite{mescheder2019occupancy} & 4.3 & 34.0 & 80 & 33 & 69 \\
$L_{Sil}$ & 12.2 & 154.8 & 51 & 16 & 87 \\
$L_{Sil, Dis, Nc, Lp}$ & 3.7 & 26.2 & 80 & 31 & 85 \\
$L_{Sil, Dis, Nc, Lp, Vsym}$ & 3.7 & 25.8 & 81 & 32 & 86 \\
\bm{$L_{total}$} & \textbf{3.3} & \textbf{22.5} & \textbf{84} & \textbf{35} & 85 \\
\hline
OccNet* \cite{mescheder2019occupancy} & 11.0 & 123.3 & 48 & 10 & 53\\
$L_{total}$, $\lambda_{SymB}=1.0$* & 8.9 & 89.1 & 52 & 10 & 72 \\
$\bm{L_{total}*}$ & \textbf{7.8} & \textbf{85.9 }& \textbf{55} & \textbf{12} & 76 \\
\hline
\end{tabular}
\end{center}
\caption{Ablation study of REFINE. $L_{total}$ indicates that all losses are used; an asterisk indicates results averaged over ShapeNetAsym instead of RerenderedShapeNet.}
\label{tab:ablation}
\end{table}

\begin{figure}
\begin{center}
\includegraphics[width=7.9cm]{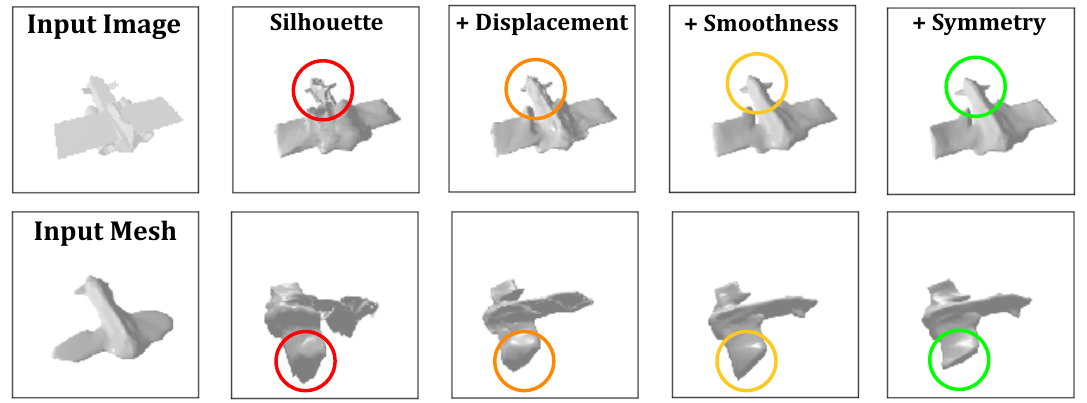} \\
\end{center}
   \caption{Leftmost column: input image and mesh. Other columns: REFINEment improves with an increasingly larger set of losses (left to right). Best viewed enlarged.}
\label{fig:loss_study}
\end{figure}

\subsection{Ablation Studies} \label{sec:results_multiview}
Ablation studies were performed for different components of REFINE. In addition to the metrics above, we also measure in these experiments the consistency between reconstructed mesh and input silhouette, by computing the 2D IoU between the latter and the silhouette of the mesh render.

The top section of Table \ref{tab:ablation} shows the effect of different REFINEments of RerenderedShapeNet meshes originally reconstructed by OccNet. The first row has not been REFINEd. The second row shows that, using the silhouette loss only ($\lambda_{Sil}=10$, all other $\lambda$s set to 0) improves input image consistency (from 69 to 87 2D IoU), but the refined mesh severely overfits to the input viewpoint, leading to decreased 3D accuracy. Adding the popular regularizers in the literature  ($\lambda_{Dis}=100$, $\lambda_{Nc}=10$, $\lambda_{Lp}=10$), limits vertex displacements and encourages mesh smoothness. As shown in the third row, this improves 3D reconstruction performance, but the gains over the baseline are small. The fourth row shows that enforcing vertex symmetry ($\lambda_{Vsym}=20, \lambda_{SymB}=0.0005$) has marginal improvements by itself. However, when combined with render-based image symmetry (row five, which further adds the image symmetry loss with $\lambda_{Isym}=100$)  it enables significant improvements in all metrics (e.g. from 34 to 22.5 Chamfer Distance). \looseness=-1

\begin{table}
\begin{center}
\fontsize{6.0}{7.5}\selectfont
\setlength\tabcolsep{4pt}
\begin{tabular}{|c|c |c| c |c |c |}
\hline
& & EMD $\downarrow$ & CD-$l_2 \downarrow$ & F-Score $\uparrow$ & Vol. IoU $\uparrow$ \\
\hline
\hline
\multirow{4}{*}{SVR} & AtlasNet \cite{groueix2018papier} & 8.0  & 13.0  & 89  & 30 \\
& Mesh R-CNN \cite{gkioxari2019mesh} & 4.2  & 10.3  & 90  & - \\
& Pix2Mesh \cite{wang2018pixel2mesh} & 3.4  & 8.0  & 93  & 48 \\
& DISN \cite{xu2019disn} & 2.6  & 9.7  & 91  & 57 \\
\Xhline{3\arrayrulewidth}
\multirow{4}{*}{TTSR} & \multirow{2}{*}{MeshSDF \cite{remelli2020meshsdf}} & 3.0$\rightarrow$2.5  & 12.0$\rightarrow$7.8  & 91$\rightarrow$95  & \multirow{2}{*}{-} \\
& & (-0.5) & (-4.2) & (+4) & \\
\cline{2-6}
&  \multirow{2}{*}{REFINEd OccNet \cite{mescheder2019occupancy}} & 2.9$\rightarrow$\textbf{2.3} & 12.2$\rightarrow$\textbf{7.5}  & 91$\rightarrow$\textbf{96}  & 57$\rightarrow$\textbf{59} \\
& & (-0.6) & (-4.7) & (+5) & (+2) \\
\hline
\end{tabular}
\end{center}
\caption{Reconstruction accuracies with no domain shift. Top: single view reconstruction (SVR) networks. Bottom: test-time shape refinement (TTSR) methods. TTSR results presented by accuracy  {\it before\/}  $\rightarrow$ {\it after\/} refinement, with gain shown in parenthesis. 
}
\label{tab:no_domain_gap}
\end{table}

\begin{table*}
\begin{center}
\fontsize{6.0}{7.5}\selectfont
\setlength\tabcolsep{5pt}
\begin{tabular}{|c?c|c|c|c?c|c|c|c?c|c|c|c|}
\hline
 & \multicolumn{4}{c?}{REFINEd OccNet \cite{mescheder2019occupancy}} & \multicolumn{4}{c?}{REFINEd Pix2Mesh \cite{wang2018pixel2mesh}} & \multicolumn{4}{c|}{REFINEd AtlasNet \cite{groueix2018papier}} \\
\hline
 & EMD $\downarrow$ & CD-$l_2 \downarrow$ & F-Score $\uparrow$ & Vol. IoU $\uparrow$ & EMD $\downarrow$ & CD-$l_2 \downarrow$ & F-Score $\uparrow$ & Vol. IoU $\uparrow$  & EMD $\downarrow$ & CD-$l_2 \downarrow$ & F-Score $\uparrow$ & Vol. IoU $\uparrow$ \\
\hline\hline
Airplane & 3.5 $\rightarrow$\textbf{ 2.2 }& 20.6 $\rightarrow$\textbf{ 11.4 }& 86 $\rightarrow$\textbf{ 91 }& 38 $\rightarrow$\textbf{ 40 }& 3.7 $\rightarrow$\textbf{ 2.3 }& 22.3 $\rightarrow$\textbf{ 11.0 }& 65 $\rightarrow$\textbf{ 88 }& 12 $\rightarrow$\textbf{ 22 }& 5.3 $\rightarrow$\textbf{ 3.8 }& 41.9 $\rightarrow$\textbf{ 18.2 }& 60 $\rightarrow$\textbf{ 82 }& 5 $\rightarrow$\textbf{ 13} \\
Bench & 2.9 $\rightarrow$\textbf{ 2.2 }& 28.6 $\rightarrow$\textbf{ 17.0 }& 84 $\rightarrow$\textbf{ 86 }& 20 $\rightarrow$\textbf{ 20 }& 3.6 $\rightarrow$\textbf{ 2.6 }& 28.0 $\rightarrow$\textbf{ 19.9 }& 65 $\rightarrow$\textbf{ 76 }& 9 $\rightarrow$\textbf{ 11 }& 4.9 $\rightarrow$\textbf{ 4.6 }& 50.0 $\rightarrow$\textbf{ 37.7 }& 58 $\rightarrow$\textbf{ 68 }& 5 $\rightarrow$\textbf{ 8} \\
Cabinet & 3.4 $\rightarrow$\textbf{ 2.7 }& 17.0 $\rightarrow$\textbf{ 14.8 }& 83 $\rightarrow$\textbf{ 85 }& 45 $\rightarrow$\textbf{ 46 }& 3.6 $\rightarrow$\textbf{ 3.0 }& 20.2 $\rightarrow$\textbf{ 16.4 }& 74 $\rightarrow$\textbf{ 78 }& 37 $\rightarrow$\textbf{ 39 }& 4.3 $\rightarrow$\textbf{ 4.1 }& 30.7 $\rightarrow$\textbf{ 19.9 }& 59 $\rightarrow$\textbf{ 75 }& 14 $\rightarrow$\textbf{ 17} \\
Car & 2.9 $\rightarrow$\textbf{ 2.5 }& 19.9 $\rightarrow$\textbf{ 12.9 }& 86 $\rightarrow$\textbf{ 87 }& 30 $\rightarrow$\textbf{ 31 }& 2.7 $\rightarrow$\textbf{ 2.3 }& 10.8 $\rightarrow$\textbf{ 7.8 }& 85 $\rightarrow$\textbf{ 90 }& 24 $\rightarrow$\textbf{ 27 }& 7.6 $\rightarrow$\textbf{ 4.8 }& 98.8 $\rightarrow$\textbf{ 27.0 }& 44 $\rightarrow$\textbf{ 72 }& 6 $\rightarrow$\textbf{ 12} \\
Chair & 6.5 $\rightarrow$\textbf{ 5.4 }& 48.5 $\rightarrow$\textbf{ 39.4 }& 72 $\rightarrow$\textbf{ 76 }& 29 $\rightarrow$\textbf{ 32 }& 6.3 $\rightarrow$\textbf{ 4.5 }& 35.4 $\rightarrow$\textbf{ 25.2 }& 60 $\rightarrow$\textbf{ 73 }& 17 $\rightarrow$\textbf{ 22 }& 6.8 $\rightarrow$\textbf{ 5.0 }& 49.5 $\rightarrow$\textbf{ 27.3 }& 53 $\rightarrow$\textbf{ 71 }& 8 $\rightarrow$\textbf{ 13} \\
Display & 3.5 $\rightarrow$\textbf{ 2.7 }& 30.8 $\rightarrow$\textbf{ 18.1 }& 76 $\rightarrow$\textbf{ 83 }& 31 $\rightarrow$\textbf{ 37 }& 4.2 $\rightarrow$\textbf{ 3.0 }& 28.0 $\rightarrow$\textbf{ 17.4 }& 72 $\rightarrow$\textbf{ 81 }& 25 $\rightarrow$\textbf{ 32 }& 4.9 $\rightarrow$\textbf{ 4.5 }& 43.1 $\rightarrow$\textbf{ 30.0 }& 61 $\rightarrow$\textbf{ 71 }& 10 $\rightarrow$\textbf{ 14} \\
Lamp & 8.9 $\rightarrow$\textbf{ 6.3 }& 90.5 $\rightarrow$\textbf{ 59.1 }& 68 $\rightarrow$\textbf{ 73 }& 22 $\rightarrow$\textbf{ 23 }& 9.2 $\rightarrow$\textbf{ 7.0 }& 71.6 $\rightarrow$\textbf{ 40.6 }& 50 $\rightarrow$\textbf{ 66 }& 11 $\rightarrow$\textbf{ 14 }& 10.2 $\rightarrow$\textbf{ 7.5 }& 102.4 $\rightarrow$\textbf{ 51.1 }& 44 $\rightarrow$\textbf{ 62 }& 5 $\rightarrow$\textbf{ 10} \\
Speakers & 4.4 $\rightarrow$\textbf{ 3.6 }& 29.8 $\rightarrow$\textbf{ 22.3 }& 73 $\rightarrow$\textbf{ 76 }& 43 $\rightarrow$\textbf{ 44 }& 4.3 $\rightarrow$\textbf{ 3.8 }& 31.4 $\rightarrow$\textbf{ 25.5 }& 65 $\rightarrow$\textbf{ 70 }& 36 $\rightarrow$\textbf{ 38 }& 5.4 $\rightarrow$\textbf{ 4.7 }& 46.6 $\rightarrow$\textbf{ 27.7 }& 55 $\rightarrow$\textbf{ 69 }& 13 $\rightarrow$\textbf{ 17} \\
Rifle & 6.5 $\rightarrow$\textbf{ 3.9 }& 37.7 $\rightarrow$\textbf{ 14.6 }& 86 $\rightarrow$\textbf{ 91 }& 30 $\rightarrow$\textbf{ 30 }& 3.5 $\rightarrow$\textbf{ 3.4 }& 18.1 $\rightarrow$\textbf{ 10.1 }& 76 $\rightarrow$\textbf{ 91 }& 12 $\rightarrow$\textbf{ 21 }& 6.3 $\rightarrow$\textbf{ 4.5 }& 61.4 $\rightarrow$\textbf{ 28.6 }& 70 $\rightarrow$\textbf{ 84 }& 7 $\rightarrow$\textbf{ 14} \\
Sofa & 3.0 $\rightarrow$\textbf{ 2.7 }& 23.8 $\rightarrow$\textbf{ 17.9 }& 83 $\rightarrow$\textbf{ 85 }& 48 $\rightarrow$\textbf{ 49 }& 4.3 $\rightarrow$\textbf{ 3.2 }& 24.8 $\rightarrow$\textbf{ 21.8 }& 71 $\rightarrow$\textbf{ 79 }& 34 $\rightarrow$\textbf{ 40 }& 5.3 $\rightarrow$\textbf{ 4.7 }& 48.0 $\rightarrow$\textbf{ 31.1 }& 63 $\rightarrow$\textbf{ 73 }& 15 $\rightarrow$\textbf{ 19} \\
Table & 4.5 $\rightarrow$\textbf{ 3.9 }& 40.6 $\rightarrow$\textbf{ 34.3 }& 72 $\rightarrow$\textbf{ 77 }& 17 $\rightarrow$\textbf{ 20 }& 9.3 $\rightarrow$\textbf{ 6.2 }& 159.3 $\rightarrow$\textbf{ 81.8 }& 30 $\rightarrow$\textbf{ 44 }& 6 $\rightarrow$\textbf{ 8 }& 8.9 $\rightarrow$\textbf{ 7.4 }& 129.6 $\rightarrow$\textbf{ 82.7 }& 36 $\rightarrow$\textbf{ 47 }& 4 $\rightarrow$\textbf{ 8} \\
Telephone & 2.3 $\rightarrow$\textbf{ 2.0 }& 10.9 $\rightarrow$\textbf{ 8.0 }& 90 $\rightarrow$\textbf{ 92 }& 48 $\rightarrow$\textbf{ 50 }& 2.2 $\rightarrow$\textbf{ 1.8 }& 10.9 $\rightarrow$\textbf{ 8.2 }& 89 $\rightarrow$\textbf{ 92 }& 40 $\rightarrow$\textbf{ 44 }& 3.4 $\rightarrow$\textbf{ 3.3 }& 33.6 $\rightarrow$\textbf{ 20.8 }& 66 $\rightarrow$\textbf{ 79 }& 11 $\rightarrow$\textbf{ 16} \\
Watercraft & 4.3 $\rightarrow$\textbf{ 2.9 }& 42.4 $\rightarrow$\textbf{ 23.5 }& 80 $\rightarrow$\textbf{ 86 }& 32 $\rightarrow$\textbf{ 36 }& 5.0 $\rightarrow$\textbf{ 2.7 }& 32.7 $\rightarrow$\textbf{ 14.0 }& 71 $\rightarrow$\textbf{ 86 }& 16 $\rightarrow$\textbf{ 27 }& 7.1 $\rightarrow$\textbf{ 4.3 }& 76.8 $\rightarrow$\textbf{ 25.5 }& 55 $\rightarrow$\textbf{ 79 }& 6 $\rightarrow$\textbf{ 15} \\
\hline
\multirow{2}{*}{Mean} & 4.3 $\rightarrow$\textbf{ 3.3 }& 34.0 $\rightarrow$\textbf{ 22.5 }& 80 $\rightarrow$\textbf{ 84 }& 33 $\rightarrow$\textbf{ 35 }& 4.8 $\rightarrow$\textbf{ 3.5 }& 38.0 $\rightarrow$\textbf{ 23.1 }& 67 $\rightarrow$\textbf{ 78 }& 22 $\rightarrow$\textbf{ 27 }& 6.2 $\rightarrow$\textbf{ 4.9 }& 62.5 $\rightarrow$\textbf{ 32.9 }& 56 $\rightarrow$\textbf{ 72 }& 8 $\rightarrow$\textbf{ 13} \\
 & (\textbf{-1.0}) & (\textbf{-11.5}) & (\textbf{+4}) & (\textbf{+2}) & (\textbf{-1.3}) & (\textbf{-14.9}) & (\textbf{+11}) & (\textbf{+5}) & (\textbf{-1.3}) & (\textbf{-29.6}) & (\textbf{+16}) & (\textbf{+5}) \\
\hline
\end{tabular}
\end{center}
\caption{REFINEment in the presence of mild domain shift, namely RerenderedShapeNet reconstructions by ShapeNet trained networks. REFINE achieves gains under all networks, classes, and metrics.
}
\label{tab:shapenet_given_pose}
\end{table*}

\begin{table}
\begin{center}
\fontsize{6.0}{7.5}\selectfont
\setlength\tabcolsep{3pt}
\begin{tabular}{|c|c|c|c|c|c|}
\hline
& & EMD $\downarrow$ & CD-$l_2 \downarrow$ & F-Score $\uparrow$ & Vol. IoU $\uparrow$ \\
\hline\hline
\multirow{2}{*}{MeshSDF \cite{remelli2020meshsdf}} & \multirow{2}{*}{Chair}  & 11.9 $\rightarrow$ 9.8 & 102.0 $\rightarrow$ 89.0 & \multirow{2}{*}{-}  & \multirow{2}{*}{-} \\
 &  & (-2.1) & (-13.0) & & \\
\hline
\multirow{10}{*}{\shortstack{REFINEd \\ OccNet \cite{mescheder2019occupancy}}} & Chair & 11.0 $\rightarrow$\textbf{ 8.5 }& 110.7 $\rightarrow$\textbf{ 74.5 }& 57 $\rightarrow$\textbf{ 62 }& 18 $\rightarrow$\textbf{ 20} \\
& Bed* & 7.5 $\rightarrow$\textbf{ 6.1 }& 70.1 $\rightarrow$\textbf{ 47.9 }& 57 $\rightarrow$\textbf{ 62 }& 22 $\rightarrow$\textbf{ 23} \\
& Bookcase* & 7.4 $\rightarrow$\textbf{ 4.1 }& 72.0 $\rightarrow$\textbf{ 38.5 }& 56 $\rightarrow$\textbf{ 65 }& 9 $\rightarrow$\textbf{ 12} \\
& Desk & 7.6 $\rightarrow$\textbf{ 6.7 }& 60.6 $\rightarrow$\textbf{ 43.7 }& 71 $\rightarrow$\textbf{ 72 }& 26 $\rightarrow$\textbf{ 27} \\
& Misc* & 10.2 $\rightarrow$\textbf{ 5.4 }& 129.6 $\rightarrow$\textbf{ 69.8 }& 46 $\rightarrow$\textbf{ 55 }& 19 $\rightarrow$\textbf{ 20} \\
& Sofa & 3.2 $\rightarrow$\textbf{ 3.1 }& 30.8 $\rightarrow$\textbf{ 25.5 }& 75 $\rightarrow$\textbf{ 76 }& 50 $\rightarrow$\textbf{ 51} \\
& Table & 6.5 $\rightarrow$\textbf{ 5.6 }& 67.7 $\rightarrow$\textbf{ 57.8 }& 60 $\rightarrow$\textbf{ 62 }& 16 $\rightarrow$\textbf{ 17} \\
& Tool* & 10.8 $\rightarrow$\textbf{ 8.6 }& 140.8 $\rightarrow$\textbf{ 118.6 }& 51 $\rightarrow$\textbf{ 60 }& 11 $\rightarrow$\textbf{ 14} \\
& Wardrobe* & 5.9 $\rightarrow$\textbf{ 3.7 }& 49.9 $\rightarrow$\textbf{ 29.3 }& 65 $\rightarrow$\textbf{ 68 }& 54 $\rightarrow$\textbf{ 55} \\
\cline{2-6}
& \multirow{2}{*}{Mean} & 7.9 $\rightarrow$\textbf{ 5.8 }& 81.4 $\rightarrow$\textbf{ 56.2 }& 59 $\rightarrow$\textbf{ 65 }& 23 $\rightarrow$\textbf{ 28} \\
&  & (\textbf{-2.1}) & (\textbf{-25.2}) & (\textbf{+6}) & (\textbf{+5}) \\
\hline
\end{tabular}
\end{center}
\caption{REFINEment gain in the presence of large domain shift, namely Pix3D reconstructions by ShapeNet trained networks. 
REFINE achieves gains under all metrics and for all networks. REFINE is even able to improve on classes not seen during training, shown with an asterisk.}
\label{tab:pix3d_given_pose}
\end{table}

The bottom three rows of Table \ref{tab:ablation} use ShapeNetAsym to study the effect of asymmetry on REFINE performance. The sixth row is not REFINEd. The seventh row shows that when the confidence scores of~(\ref{eq:vertex_sym}) and (\ref{eq:render_sym}) are removed (by setting $\lambda_{SymB} = 1$, in which case the confidence scores become approximately 1) the refinement of asymmetrical meshes is significantly less accurate than that of the default configuration (eighth row, $\lambda_{SymB}=0.0005$). It can also be seen that, when confidence scores are used, the reconstruction quality is significantly superior to that of the original meshes. In summary, the proposed confidence mechanism enables effective REFINEment of non-symmetric objects.

Figure \ref{fig:loss_study} illustrates the contribution of each loss. The leftmost column shows the input airplane image (top) and mesh (bottom). From the second column, we progressively add more losses. With only the silhouette loss, REFINE severely overfits to the input viewpoint. The displacement loss helps regularize deformation magnitude; the smoothness losses reduce jagged artifacts; the symmetry losses correct shape details (e.g. airplane tail) by enforcing a symmetry prior. These operate intuitively and can be tweaked for target applications. For example, if only symmetric objects will be reconstructed $\lambda_{SymB}$ can be increased.

\subsection{Refinement Results} \label{sec:results}
We next consider the robustness of REFINE postprocessing to different levels of domain gap. First, refinement experiments without domain gap were performed, for reconstruction networks trained and tested on the ShapeNet renders of \cite{choy20163d}. Table \ref{tab:no_domain_gap} presents results for different reconstruction networks and test-time refinement methods (please see supplementary for full per-class results). As the particulars of MeshSDF's architecture~\cite{remelli2020meshsdf} have not been open sourced as of this submission, we instead REFINE OccNet~\cite{mescheder2019occupancy}; here, OccNet and the raw, unrefined version of MeshSDF are directly comparable as both are implicit based, with nearly identical performance prior to refinement. However, REFINE achieves state of the art test-time shape refinement results when paired with OccNet, despite being a black-box method compared to the white-box refinement of MeshSDF~\cite{remelli2020meshsdf}, which is specific to their network.


Several experiments were next conducted to evaluate the effectiveness of REFINEment in the presence of domain gap. Table \ref{tab:shapenet_given_pose} gives reconstruction accuracy for RerenderedShapeNet reconstructions, before and after REFINEment, of ShapeNet pretrained networks. Three methods representative of different reconstruction strategies are considered: OccNet \cite{mescheder2019occupancy}, based on implicit functions, Pixel2Mesh \cite{wang2018pixel2mesh}, which deforms an ellipsoid, and AtlasNet \cite{groueix2018papier}, based on surface atlas elements. A larger table with more REFINEd methods is presented in the supplementary. The results of Table \ref{tab:shapenet_given_pose} are generally worse than those of Table \ref{tab:no_domain_gap}; while the methods perform well on the training domain, they struggle to generalize to out-of-distribution data. However, REFINEment significantly recovers much of the degraded performance for all networks. Average gains are particularly large under the Chamfer distance (-11.5 for OccNet, -14.9 for Pixel2Mesh, and -29.6 for AtlasNet) and increase with the  network sensitivity to the domain gap (e.g. largest for AtlasNet, which has the weakest performance). 

Finally, the effectiveness of REFINE is validated on the real-world Pix3D dataset, which has the largest domain gap. Table \ref{tab:pix3d_given_pose} shows that the beneficial behavior of REFINE is qualitatively identical to that of Table~\ref{tab:shapenet_given_pose}. A comparison to the test-time refinement method of~\cite{remelli2020meshsdf} on the shapes of the ``Chair'' class (the only for which \cite{remelli2020meshsdf} reports results) once again shows that REFINE substantially improves on the state of the art. This occurs even though the initial performance prior to refinement is actually worse for OccNet compared to MeshSDF in this case (e.g. Chamfer Distance 110.7 versus 102), demonstrating the strength of REFINE.

Overall, Tables \ref{tab:no_domain_gap}, \ref{tab:shapenet_given_pose}, and \ref{tab:pix3d_given_pose} show that REFINE postprocessing achieves state of the art reconstruction accuracies,  consistently improving the performance of original reconstruction networks regardless of class, metric, base reconstruction method, or dataset. Furthermore, gains increase with the domain gap; for the best performing OccNet (ShapeNet trained), REFINE yields Chamfer Distance average improvements of -4.7, -11.5, and -25.2 on ShapeNet, RerenderedShapeNet, and Pix3D respectively.

These improvements are illustrated by examples in Figure \ref{fig:qual_examples} (originally reconstructed by an OccNet). REFINE can both sharpen details (such as the airplane's elongated nose) and create entirely new parts (set of wings in the back). It also excels at recovering details from unusual ``outlier'' shapes, such as cars with spoilers and is successful even  for {\it classes on which the original reconstruction method was not trained\/}, leading to poor original meshes. These are marked by an asterisk in Table \ref{tab:pix3d_given_pose}, and include the bed and spoon in the first two rows of Figure \ref{fig:qual_examples}. REFINE can also recover from very poor reconstructions due to significant domain shift, such as the Pix3D table shown in the third row of Figure \ref{fig:qual_examples}. Finally, Figure \ref{fig:method_compare} illustrates that although OccNet, Pix2Mesh, and AtlasNet produce very different failure cases and artifacts, REFINE improves the both the input image consistency and 3D accuracy of all methods. Additional examples can be found in the supplementary.

\section{Conclusion}
In this paper, we demonstrated the effectiveness and versatility of black-box mesh refinement at test time for the problem of single view 3D reconstruction. The proposed REFINE method enforces regularized input image consistency, and can be applied to any reconstruction method in the literature. Experiments show systematic significant improvements over the state of the art, for many metrics, datasets, and reconstruction methods. We believe that this new paradigm will remain relevant as novel reconstruction networks are introduced, and inspire substantial future work in test-time black-box refinement of reconstructed meshes.

{\small
\bibliographystyle{ieee_fullname}
\bibliography{egbib}

\begin{thebibliography}{10}\itemsep=-1pt

\bibitem{bernardini1999ball}
Fausto Bernardini, Joshua Mittleman, Holly Rushmeier, Cl{\'a}udio Silva, and
  Gabriel Taubin.
\newblock The ball-pivoting algorithm for surface reconstruction.
\newblock {\em IEEE transactions on visualization and computer graphics},
  5(4):349--359, 1999.

\bibitem{calakli2011ssd}
Fatih Calakli and Gabriel Taubin.
\newblock Ssd: Smooth signed distance surface reconstruction.
\newblock In {\em Computer Graphics Forum}, volume~30, pages 1993--2002. Wiley
  Online Library, 2011.

\bibitem{chang2015shapenet}
Angel~X Chang, Thomas Funkhouser, Leonidas Guibas, Pat Hanrahan, Qixing Huang,
  Zimo Li, Silvio Savarese, Manolis Savva, Shuran Song, Hao Su, et~al.
\newblock Shapenet: An information-rich 3d model repository.
\newblock {\em arXiv preprint arXiv:1512.03012}, 2015.

\bibitem{choy20163d}
Christopher~B Choy, Danfei Xu, JunYoung Gwak, Kevin Chen, and Silvio Savarese.
\newblock 3d-r2n2: A unified approach for single and multi-view 3d object
  reconstruction.
\newblock In {\em European conference on computer vision}, pages 628--644.
  Springer, 2016.

\bibitem{blender}
Blender~Online Community.
\newblock {\em Blender - a 3D modelling and rendering package}.
\newblock Blender Foundation, Stichting Blender Foundation, Amsterdam, 2018.

\bibitem{desbrun1999implicit}
Mathieu Desbrun, Mark Meyer, Peter Schr{\"o}der, and Alan~H Barr.
\newblock Implicit fairing of irregular meshes using diffusion and curvature
  flow.
\newblock In {\em Proceedings of the 26th annual conference on Computer
  graphics and interactive techniques}, pages 317--324, 1999.

\bibitem{fan2017point}
Haoqiang Fan, Hao Su, and Leonidas~J Guibas.
\newblock A point set generation network for 3d object reconstruction from a
  single image.
\newblock In {\em Proceedings of the IEEE conference on computer vision and
  pattern recognition}, pages 605--613, 2017.

\bibitem{ganin2016domain}
Yaroslav Ganin, Evgeniya Ustinova, Hana Ajakan, Pascal Germain, Hugo
  Larochelle, Fran{\c{c}}ois Laviolette, Mario Marchand, and Victor Lempitsky.
\newblock Domain-adversarial training of neural networks.
\newblock {\em The Journal of Machine Learning Research}, 17(1):2096--2030,
  2016.

\bibitem{gao2019prs}
Lin Gao, Ling-Xiao Zhang, Hsien-Yu Meng, Yi-Hui Ren, Yu-Kun Lai, and Leif
  Kobbelt.
\newblock Prs-net: Planar reflective symmetry detection net for 3d models.
\newblock {\em arXiv preprint arXiv:1910.06511}, 2019.

\bibitem{genova2020local}
Kyle Genova, Forrester Cole, Avneesh Sud, Aaron Sarna, and Thomas Funkhouser.
\newblock Local deep implicit functions for 3d shape.
\newblock In {\em Proceedings of the IEEE/CVF Conference on Computer Vision and
  Pattern Recognition}, pages 4857--4866, 2020.

\bibitem{gkioxari2019mesh}
Georgia Gkioxari, Jitendra Malik, and Justin Johnson.
\newblock Mesh r-cnn.
\newblock In {\em Proceedings of the IEEE International Conference on Computer
  Vision}, pages 9785--9795, 2019.

\bibitem{groueix2018papier}
Thibault Groueix, Matthew Fisher, Vladimir~G Kim, Bryan~C Russell, and Mathieu
  Aubry.
\newblock A papier-m{\^a}ch{\'e} approach to learning 3d surface generation.
\newblock In {\em Proceedings of the IEEE conference on computer vision and
  pattern recognition}, pages 216--224, 2018.

\bibitem{he2016deep}
Kaiming He, Xiangyu Zhang, Shaoqing Ren, and Jian Sun.
\newblock Deep residual learning for image recognition.
\newblock In {\em Proceedings of the IEEE conference on computer vision and
  pattern recognition}, pages 770--778, 2016.

\bibitem{jang2019interactive}
Won-Dong Jang and Chang-Su Kim.
\newblock Interactive image segmentation via backpropagating refinement scheme.
\newblock In {\em Proceedings of the IEEE Conference on Computer Vision and
  Pattern Recognition}, pages 5297--5306, 2019.

\bibitem{kanazawa2018learning}
Angjoo Kanazawa, Shubham Tulsiani, Alexei~A Efros, and Jitendra Malik.
\newblock Learning category-specific mesh reconstruction from image
  collections.
\newblock In {\em Proceedings of the European Conference on Computer Vision
  (ECCV)}, pages 371--386, 2018.

\bibitem{kato2018neural}
Hiroharu Kato, Yoshitaka Ushiku, and Tatsuya Harada.
\newblock Neural 3d mesh renderer.
\newblock In {\em Proceedings of the IEEE Conference on Computer Vision and
  Pattern Recognition}, pages 3907--3916, 2018.

\bibitem{kazhdan2006poisson}
Michael Kazhdan, Matthew Bolitho, and Hugues Hoppe.
\newblock Poisson surface reconstruction.
\newblock In {\em Proceedings of the fourth Eurographics symposium on Geometry
  processing}, volume~7, 2006.

\bibitem{kazhdan2013screened}
Michael Kazhdan and Hugues Hoppe.
\newblock Screened poisson surface reconstruction.
\newblock {\em ACM Transactions on Graphics (ToG)}, 32(3):1--13, 2013.

\bibitem{kipf2016semi}
Thomas~N Kipf and Max Welling.
\newblock Semi-supervised classification with graph convolutional networks.
\newblock {\em arXiv preprint arXiv:1609.02907}, 2016.

\bibitem{li2020onlineAda}
Xueting Li, Sifei Liu, Shalini De~Mello, Kihwan Kim, Xiaolong Wang, Ming-Hsuan
  Yang, and Jan Kautz.
\newblock Online adaptation for consistent mesh reconstruction in the wild.
\newblock In {\em Advances in Neural Information Processing Systems}, 2020.

\bibitem{li2020self}
Xueting Li, Sifei Liu, Kihwan Kim, Shalini De~Mello, Varun Jampani, Ming-Hsuan
  Yang, and Jan Kautz.
\newblock Self-supervised single-view 3d reconstruction via semantic
  consistency.
\newblock {\em arXiv preprint arXiv:2003.06473}, 2020.

\bibitem{lin2017learning}
Chen-Hsuan Lin, Chen Kong, and Simon Lucey.
\newblock Learning efficient point cloud generation for dense 3d object
  reconstruction.
\newblock {\em arXiv preprint arXiv:1706.07036}, 2017.

\bibitem{lin2019photometric}
Chen-Hsuan Lin, Oliver Wang, Bryan~C Russell, Eli Shechtman, Vladimir~G Kim,
  Matthew Fisher, and Simon Lucey.
\newblock Photometric mesh optimization for video-aligned 3d object
  reconstruction.
\newblock In {\em Proceedings of the IEEE Conference on Computer Vision and
  Pattern Recognition}, pages 969--978, 2019.

\bibitem{liu2019soft}
Shichen Liu, Tianye Li, Weikai Chen, and Hao Li.
\newblock Soft rasterizer: A differentiable renderer for image-based 3d
  reasoning.
\newblock In {\em Proceedings of the IEEE International Conference on Computer
  Vision}, pages 7708--7717, 2019.

\bibitem{liu2010computational}
Yanxi Liu, Hagit Hel-Or, and Craig~S Kaplan.
\newblock {\em Computational symmetry in computer vision and computer
  graphics}.
\newblock Now publishers Inc, 2010.

\bibitem{long2017deep}
Mingsheng Long, Han Zhu, Jianmin Wang, and Michael~I Jordan.
\newblock Deep transfer learning with joint adaptation networks.
\newblock In {\em International conference on machine learning}, pages
  2208--2217. PMLR, 2017.

\bibitem{lorensen1987marching}
William~E Lorensen and Harvey~E Cline.
\newblock Marching cubes: A high resolution 3d surface construction algorithm.
\newblock {\em ACM siggraph computer graphics}, 21(4):163--169, 1987.

\bibitem{mescheder2019occupancy}
Lars Mescheder, Michael Oechsle, Michael Niemeyer, Sebastian Nowozin, and
  Andreas Geiger.
\newblock Occupancy networks: Learning 3d reconstruction in function space.
\newblock In {\em Proceedings of the IEEE Conference on Computer Vision and
  Pattern Recognition}, pages 4460--4470, 2019.

\bibitem{michalkiewicz2020few}
Mateusz Michalkiewicz, Sarah Parisot, Stavros Tsogkas, Mahsa Baktashmotlagh,
  Anders Eriksson, and Eugene Belilovsky.
\newblock Few-shot single-view 3-d object reconstruction with compositional
  priors.
\newblock {\em arXiv preprint arXiv:2004.06302}, 2020.

\bibitem{niemeyer2020differentiable}
Michael Niemeyer, Lars Mescheder, Michael Oechsle, and Andreas Geiger.
\newblock Differentiable volumetric rendering: Learning implicit 3d
  representations without 3d supervision.
\newblock In {\em Proceedings of the IEEE/CVF Conference on Computer Vision and
  Pattern Recognition}, pages 3504--3515, 2020.

\bibitem{park2019deepsdf}
Jeong~Joon Park, Peter Florence, Julian Straub, Richard Newcombe, and Steven
  Lovegrove.
\newblock Deepsdf: Learning continuous signed distance functions for shape
  representation.
\newblock In {\em Proceedings of the IEEE Conference on Computer Vision and
  Pattern Recognition}, pages 165--174, 2019.

\bibitem{pinheiro2019domain}
Pedro~O Pinheiro, Negar Rostamzadeh, and Sungjin Ahn.
\newblock Domain-adaptive single-view 3d reconstruction.
\newblock In {\em Proceedings of the IEEE International Conference on Computer
  Vision}, pages 7638--7647, 2019.

\bibitem{ravi2020pytorch3d}
Nikhila Ravi, Jeremy Reizenstein, David Novotny, Taylor Gordon, Wan-Yen Lo,
  Justin Johnson, and Georgia Gkioxari.
\newblock Accelerating 3d deep learning with pytorch3d.
\newblock {\em arXiv:2007.08501}, 2020.

\bibitem{remelli2020meshsdf}
Edoardo Remelli, Artem Lukoianov, Stephan~R Richter, Beno{\^\i}t Guillard,
  Timur Bagautdinov, Pierre Baque, and Pascal Fua.
\newblock Meshsdf: Differentiable iso-surface extraction.
\newblock {\em arXiv preprint arXiv:2006.03997}, 2020.

\bibitem{sakinis2019interactive}
Tomas Sakinis, Fausto Milletari, Holger Roth, Panagiotis Korfiatis, Petro
  Kostandy, Kenneth Philbrick, Zeynettin Akkus, Ziyue Xu, Daguang Xu, and
  Bradley~J Erickson.
\newblock Interactive segmentation of medical images through fully
  convolutional neural networks.
\newblock {\em arXiv preprint arXiv:1903.08205}, 2019.

\bibitem{sofiiuk2020f}
Konstantin Sofiiuk, Ilia Petrov, Olga Barinova, and Anton Konushin.
\newblock f-brs: Rethinking backpropagating refinement for interactive
  segmentation.
\newblock In {\em Proceedings of the IEEE/CVF Conference on Computer Vision and
  Pattern Recognition}, pages 8623--8632, 2020.

\bibitem{sun2016deep}
Baochen Sun and Kate Saenko.
\newblock Deep coral: Correlation alignment for deep domain adaptation.
\newblock In {\em European conference on computer vision}, pages 443--450.
  Springer, 2016.

\bibitem{sun2018pix3d}
Xingyuan Sun, Jiajun Wu, Xiuming Zhang, Zhoutong Zhang, Chengkai Zhang, Tianfan
  Xue, Joshua~B Tenenbaum, and William~T Freeman.
\newblock Pix3d: Dataset and methods for single-image 3d shape modeling.
\newblock In {\em Proceedings of the IEEE Conference on Computer Vision and
  Pattern Recognition}, pages 2974--2983, 2018.

\bibitem{sun2020test}
Yu Sun, Xiaolong Wang, Zhuang Liu, John Miller, Alexei~A Efros, and Moritz
  Hardt.
\newblock Test-time training with self-supervision for generalization under
  distribution shifts.
\newblock In {\em International Conference on Machine Learning (ICML)}, 2020.

\bibitem{tatarchenko2017octree}
Maxim Tatarchenko, Alexey Dosovitskiy, and Thomas Brox.
\newblock Octree generating networks: Efficient convolutional architectures for
  high-resolution 3d outputs.
\newblock In {\em Proceedings of the IEEE International Conference on Computer
  Vision}, pages 2088--2096, 2017.

\bibitem{tatarchenko2019single}
Maxim Tatarchenko, Stephan~R Richter, Ren{\'e} Ranftl, Zhuwen Li, Vladlen
  Koltun, and Thomas Brox.
\newblock What do single-view 3d reconstruction networks learn?
\newblock In {\em Proceedings of the IEEE Conference on Computer Vision and
  Pattern Recognition}, pages 3405--3414, 2019.

\bibitem{tung2017self}
Hsiao-Yu Tung, Hsiao-Wei Tung, Ersin Yumer, and Katerina Fragkiadaki.
\newblock Self-supervised learning of motion capture.
\newblock In {\em Advances in Neural Information Processing Systems}, pages
  5236--5246, 2017.

\bibitem{tzeng2017adversarial}
Eric Tzeng, Judy Hoffman, Kate Saenko, and Trevor Darrell.
\newblock Adversarial discriminative domain adaptation.
\newblock In {\em Proceedings of the IEEE conference on computer vision and
  pattern recognition}, pages 7167--7176, 2017.

\bibitem{wallace2019few}
Bram Wallace and Bharath Hariharan.
\newblock Few-shot generalization for single-image 3d reconstruction via
  priors.
\newblock In {\em Proceedings of the IEEE International Conference on Computer
  Vision}, pages 3818--3827, 2019.

\bibitem{wang2020fully}
Dequan Wang, Evan Shelhamer, Shaoteng Liu, Bruno Olshausen, and Trevor Darrell.
\newblock Fully test-time adaptation by entropy minimization.
\newblock {\em arXiv preprint arXiv:2006.10726}, 2020.

\bibitem{wang2018pixel2mesh}
Nanyang Wang, Yinda Zhang, Zhuwen Li, Yanwei Fu, Wei Liu, and Yu-Gang Jiang.
\newblock Pixel2mesh: Generating 3d mesh models from single rgb images.
\newblock In {\em Proceedings of the European Conference on Computer Vision
  (ECCV)}, pages 52--67, 2018.

\bibitem{wang2017cnn}
Peng-Shuai Wang, Yang Liu, Yu-Xiao Guo, Chun-Yu Sun, and Xin Tong.
\newblock O-cnn: Octree-based convolutional neural networks for 3d shape
  analysis.
\newblock {\em ACM Transactions on Graphics (TOG)}, 36(4):1--11, 2017.

\bibitem{wu2017marrnet}
Jiajun Wu, Yifan Wang, Tianfan Xue, Xingyuan Sun, Bill Freeman, and Josh
  Tenenbaum.
\newblock Marrnet: 3d shape reconstruction via 2.5 d sketches.
\newblock In {\em Advances in neural information processing systems}, pages
  540--550, 2017.

\bibitem{xie2019pix2vox}
Haozhe Xie, Hongxun Yao, Xiaoshuai Sun, Shangchen Zhou, and Shengping Zhang.
\newblock Pix2vox: Context-aware 3d reconstruction from single and multi-view
  images.
\newblock In {\em Proceedings of the IEEE International Conference on Computer
  Vision}, pages 2690--2698, 2019.

\bibitem{xu2019disn}
Qiangeng Xu, Weiyue Wang, Duygu Ceylan, Radomir Mech, and Ulrich Neumann.
\newblock Disn: Deep implicit surface network for high-quality single-view 3d
  reconstruction.
\newblock In {\em Advances in Neural Information Processing Systems}, pages
  492--502, 2019.

\bibitem{zhou2020learning}
Yichao Zhou, Shichen Liu, and Yi Ma.
\newblock Learning to detect 3d reflection symmetry for single-view
  reconstruction.
\newblock {\em arXiv preprint arXiv:2006.10042}, 2020.

\bibitem{zuffi2019three}
Silvia Zuffi, Angjoo Kanazawa, Tanya Berger-Wolf, and Michael Black.
\newblock Three-d safari: Learning to estimate zebra pose, shape, and texture
  from images “in the wild”.
\newblock In {\em 2019 IEEE/CVF International Conference on Computer Vision
  (ICCV)}, pages 5358--5367. IEEE, 2019.

\end{thebibliography}
}

\end{document}